\documentclass[10pt,journal, twocolumn]{IEEEtran}
\ifCLASSOPTIONcompsoc
\else
\fi

\ifCLASSINFOpdf
\else
\fi

\usepackage[table]{xcolor}
\usepackage{times}
\usepackage{epsfig}
\usepackage{graphicx}
\usepackage{amsmath}
\usepackage{amssymb}
\usepackage{epsfig}
\usepackage{graphicx}
\usepackage{amsmath}
\usepackage{amssymb}
\usepackage{bbm}

\usepackage{algpseudocode,algorithm,algorithmicx}

\usepackage{array,setspace, amsfonts,url,bm, cellspace}
\usepackage{tikz}
\usepackage{epstopdf}
\usetikzlibrary{positioning}
\usetikzlibrary{arrows}
\usepackage{pifont}
\usepackage{multirow, hhline}

\newcolumntype{M}[1]{>{\centering\arraybackslash}m{#1}}

\usepackage[pagebackref=true,breaklinks=true,letterpaper=true,colorlinks,bookmarks=false]{hyperref}

\begin{document}

\title{Unconstrained Biometric Recognition: Summary of Recent SOCIA Lab. Research}

\author{Varsha Balakrishnan \\ University of Beira Interior, Portugal
\IEEEcompsocitemizethanks{University of Beira Interior, R. Marqu\^es D'Avila e Bolama, 6200-001, Covilh\~{a}, Portugal, E-mail: balavarsha238@gmail.com }
\thanks{}}

\markboth{Technical Report}%
{Shell \MakeLowercase{\textit{et al.}}: Bare Demo of IEEEtran.cls for Computer Society Journals}

\IEEEtitleabstractindextext{%
\begin{abstract}
The development of biometric recognition solutions able to work in visual surveillance conditions, i.e., in unconstrained data acquisition conditions and under covert protocols has been motivating growing  efforts  from the research community.  Among the various laboratories, schools and research institutes concerned  about this problem, the SOCIA: Soft Computing and Image Analysis Lab., of the University of Beira Interior, Portugal, has been among the most active in pursuing disruptive solutions for obtaining such extremely ambitious kind of automata. This report summarises the research works published by elements of the SOCIA Lab. in the last decade in the scope of biometric recognition in unconstrained conditions. The idea is that it can be used as basis for someone wishing to entering in this research topic.
\end{abstract}

}

\maketitle

\IEEEdisplaynontitleabstractindextext

\IEEEpeerreviewmaketitle

\section{Data Acquisition Works}

\paragraph{Visible-wavelength Iris / Periocular Imaging and Recognition in Surveillance Environments~\cite{ProencaNeves2016}}

Visual surveillance cameras have been massively deployed in public urban environments over the recent years. This  raised the interest in developing automata to infer useful information from such crowded scenes (from abnormal behavior detection to human identification). In order to cover wide outdoor areas, one interesting possibility is to combine wide- angle and pan-tilt-zoom (PTZ) cameras in a master-slave configuration. The use of fish-eye lenses allows the master camera to maximize the coverage area while the PTZ acts as a foveal sensor, providing high-resolution images of the interest regions. This paper addresses the feasibility of using this type of data acquisition paradigm for imaging iris/periocular data with enough discriminating power to be used for biometric recognition purposes.

\paragraph{A Master-slave Calibration Algorithm with Fish-eye Correction~\cite{NevesProenca2015}}
Surveillance systems capable of autonomously monitoring vast areas are an emerging trend, particularly when wide-angle cameras are combined with pan-tilt-zoom (PTZ) cameras in a master-slave configuration. The use of fish-eye lenses allows the master camera to maximize the coverage area while the PTZ acts as a foveal sensor, providing high-resolution images of regions of interest. Despite the advantages of this architecture, the mapping between image coordinates and pan-tilt values is the major bottleneck in such systems, since it depends on depth information and fish-eye effect correction. In this paper, authors address these problems by exploiting geometric cues to perform height estimation. This information is used both for inferring 3D information from a single static camera deployed on an arbitrary position and for determining lens parameters to remove fish-eye distortion. When compared with the previous approaches, this method has the following advantages: (1) fish-eye distortion is corrected without relying on calibration patterns; (2) 3D information is inferred from a single static camera disposed on an arbitrary location of the scene.

\begin{figure}[h!]
  \centering
      \includegraphics[width=0.5\textwidth]{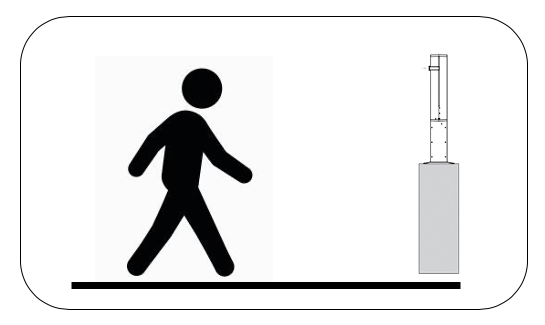}
  \caption{Schema of a non-cooperative data acquisition framework.}
\end{figure}

\paragraph{Biometric Recognition in Surveillance Environments Using Master-Slave Architectures~\cite{ProencaNeves2018}}

The number of visual surveillance systems deployed worldwide has been growing astoundingly. As a result, attempts have been made to increase the levels of automated analysis of such systems, towards the reliable recognition of human beings in fully covert conditions. Among other possibilities, master-slave architectures can be used to acquire high resolution data of subjects heads from large distances, with enough resolution to perform face recognition. This paper/tutorial provides a comprehensive overview of the major phases behind the development of a recognition system working in outdoor surveillance scenarios, describing frameworks and methods to: 1) use coupled wide view and Pan-Tilt-Zoom (PTZ) imaging devices in surveillance settings, with a wide-view camera covering the whole scene, while a synchronized PTZ device collects high-resolution data from the head region; 2) use soft biometric information (e.g., body metrology and gait) for pruning the set of potential identities for each query; and 3) faithfully balance ethics/privacy and safety/security issues in this kind of systems.

\paragraph{Acquiring High-resolution Face Images in Outdoor Environments: A master-slave Calibration Algorithm~\cite{NevesProenca2015a}}

Facial recognition at-a-distance in surveillance scenarios remains an open problem, particularly due to the small number of pixels representing the facial region. The use of pan-tilt-zoom (PTZ) cameras has been advocated to solve this problem, however, the existing approaches either rely on rough approximations or additional constraints to estimate the mapping between image coordinates and pan-tilt parameters. In this paper, authors aim at extending PTZ-assisted facial recognition to surveillance scenarios by proposing a master-slave calibration algorithm capable of accurately estimating pan-tilt parameters without depending on additional constraints. This paper exploits geometric cues to automatically estimate subjects height and thus determine their 3D position. Experimental results show that the presented algorithm is able to acquire high-resolution face im- ages at a distance ranging from 5 to 40 meters with high success rate. Additionally, authors certify the applicability of the aforementioned algorithm to biometric recognition through a face recognition test, comprising 20 probe subjects and 13,020 gallery subjects.

\paragraph{Dynamic Camera Scheduling for Visual Surveillance in Crowded Scenes using Markov Random Fields~\cite{NevesProenca2015b}}

The use of pan-tilt-zoom (PTZ) cameras for capturing high-resolution data of human-beings is an emerging trend in surveillance systems. However, this new paradigm en- tails additional challenges, such as camera scheduling, that can dramatically affect the performance of the system.  This paper presents a camera scheduling approach capable of determining - in real-time - the sequence of acquisitions that maximizes the number of different targets obtained, while minimizing the cumulative transition time. This approach models the problem as an undirected graphical model (Markov random field, MRF), which energy minimization can approximate the shortest tour to visit the maximum number of targets. A comparative analysis with the state-of-the-art camera scheduling methods evidences that this approach is able to improve the observation rate while maintaining a competitive tour time.

\paragraph{A Calibration Algorithm for Multi-camera Visual Surveillance Systems Based on Single-View Metrology~\cite{NevesProenca2015c}}

The growing concerns about persons security and the increasing popularity of pan-tilt-zoom (PTZ) cameras, have been raising the interest on automated master-slave surveillance systems. Such systems are typically composed by (1) a fixed wide-angle camera that covers a large area, detects and tracks moving objects in the scene; and (2) a PTZ camera, that provides a close-up view of an object of interest. Previously published approaches attempted to establish 2D correspondences between the video streams of both cameras, which is a ill-posed formulation due to the absence of depth information. On the other side, 3D-based approaches are more accurate but require more than one fixed camera to estimate depth information. This paper describes a novel method for easy and precise calibration of a master-slave surveillance system, composed by a single fixed wide-angle camera. The method exploits single view metrology to infer 3D data of the tracked humans and to self-perform the transformation between camera views. Experimental results in both simulated and realistic scenes point for the effectiveness of the proposed model in comparison with the state-of-the-art

\section{Databases}

\paragraph{An Annotated Multi-biometrics Data Feed From Surveillance Scenarios~\cite{NevesProenca2017b}}

The accuracy of biometric recognition in unconstrained scenarios has been a major concern for a large number of researchers. Despite such efforts, no system can recognize in a fully automated manner human beings in totally wild conditions, such as in surveillance environments. In this context, several sets of degraded data have been made available to the research community, where the reported performance by state-of-the-art algorithms is already saturated, suggesting that these sets do not reflect faithfully the conditions in such hard settings. To this end, authors introduce the QUIS-CAMPI data feed, comprising samples automatically acquired by an outdoor visual surveillance system, with subjects on-the-move and at-a-distance (up to 50 m). When compared to similar data sources, the major novelties of QUIS-CAMPI are: 1) biometric samples are acquired in a fully automatic way; 2) it is an open dataset, i.e., the number of probe images and enrolled subjects grow on a daily basis; and 3) it contains multi-biometric traits. The ensemble properties of QUIS-CAMPI ensure that the data span a representative set of covariate factors of real-world scenarios, making it a valuable tool for developing and benchmarking biometric recognition algorithms capable of working in unconstrained scenarios.

\paragraph{BioHDD: A Dataset for Studying Biometric Identification on Heavily Degraded Data~\cite{SantosProenca2015}}

This study focuses on biometric recognition in extremely degraded data, and its main contributions are three-fold: (1) announce the availability of an annotated dataset that contains high quality mugshots of 101 subjects, and large sets of probes degraded extremely by 10 different noise factors; (2) report the results of a mimicked watchlist identification scheme: an online survey was conducted, where participants were asked to perform positive and negative identification of probes against the enrolled identities. Along with their answers, volunteers had to provide the major reasons that sustained their responses, which enabled the authors to perceive the kind of features that are most frequently associated with successful/failed human identification processes. As main conclusions, the authors observed that humans rely greatly on shape information and holistic features. Otherwise, colour and texture-based features are almost disregarded by humans; (3) finally, the authors give evidence that the positive human identification on such extremely degraded data might be unreliable, whereas negative identification might constitute an interesting alternative for such cases.

\paragraph{Iris Biometrics: Synthesis of Degraded Ocular Images~\cite{CardosoProenca2015}}

Iris recognition is a popular technique for recognizing humans. However, as is the case with most biometric traits, it is difficult to collect data that are suitable for use in experiments due to three factors: 1) the substantial amount of data that is required; 2) the time that is spent in the acquisition process; and 3) the security and privacy concerns of potential volunteers. This paper describes a stochastic method for synthesizing ocular data to support experiments on iris recognition. Specifically, synthetic data are intended for use in the most important phases of those experiments: segmentation and signature encoding/matching. The resulting data have an important characteristic: they simulate image acquisition under uncontrolled conditions. authors have experimentally confirmed that the proposed strategy can mimic the data degradation factors that usually result from such conditions. Finally, authors announce the availability of an online platform for generating degraded synthetic ocular data,  freely accessible worldwide.

\paragraph{The UBIRIS.v2: A Database of Visible Wavelength Iris Images Captured On-The-Move and At-A-Distance~\cite{ProencaAlexandre2010b}}

The main purpose of this paper is to announce the availability of the UBIRIS.v2 database, a multisession iris images database which singularly contains data captured in the visible wavelength, at-a-distance (between four and eight meters) and on on-the-move. This database is freely available for researchers concerned about visible wavelength iris recognition and will be useful in accessing the feasibility and specifying the constraints of this type of biometric recognition.

\paragraph{UBEAR: A Dataset of Ear Images Captured On-the-move in Uncontrolled Conditions~\cite{RaposoProenca2011}}

In order to broad the applicability of biometric systems, the data acquisition constraints required for reliable recognition are receiving increasing attention. For some of the traits (e.g., face and iris) significant research efforts were already made toward the development of systems able to operate in completely unconstrained conditions. For other traits (e.g., the ear) no similar efforts are known. The main purpose of this paper is to announce the availability of a new data set of ear images, which main distinguishing feature is that its images were acquired from on-the-move subjects, under varying lighting conditions and without demanding to subjects any particular care regarding ear occlusions and poses. The data set is freely available to the research community and should constitute a valuable tool in assessing the possibility of performing reliable ear biometric recognition in such d challenging conditions.

\paragraph{UBIRIS: a noisy iris image database~\cite{ProencaAlexandre2005}}

This paper presents a new iris database that contains images with noise. This is in contrast with the existing databases, that are noise free. UBIRIS is a tool for the development of robust iris recognition algorithms for biometric proposes.
authors present a detailed description of the many characteristics of UBIRIS and a comparison of several image segmentation approaches used in the current iris segmentation methods where it is evident their small tolerance to noisy images.

\section{Segmentation Works}

\begin{figure}[h!]
  \centering
      \includegraphics[width=0.5\textwidth]{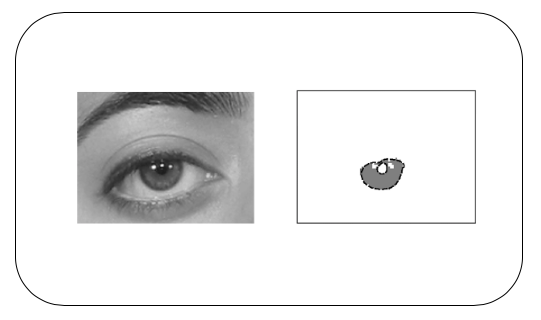}
  \caption{Iris segmentation step. receiving one close up image of the ocular region, the iris boundaries should be discriminated and parameterized.}
\end{figure}

\paragraph{Soft Biometrics: Globally Coherent Solutions for Hair Segmentation and Style Recognition based on Hierarchical MRFs\cite{ProencaNeves2017}} Markov Random Fields (MRFs) are a popular tool in many computer vision problems and faithfully model a broad range of local dependencies. However, rooted in the Hammersley-Clifford theorem, they face serious difficulties in enforcing the global coherence of the solutions without using too high order cliques that reduce the computational effectiveness of the inference phase. Having this problem in mind, authors describe a multi-layered (hierarchical) architecture for MRFs that is based exclusively in pairwise connections and typically produces globally coherent solutions, with 1) one layer working at the local (pixel) level, modelling the interactions between adjacent image patches; and 2) a complementary layer working at the object (hypothesis) level pushing toward globally consistent solutions. During optimization, both layers interact into an equilibrium state, that not only segments the data, but also classifies it. The proposed MRF architecture is particularly suitable for problems that deal with biological data (e.g., biometrics), where the reasonability of the solutions can be objectively measured.

\paragraph{Iris Recognition: On the Segmentation of Degraded Images Acquired in the Visible Wavelength~\cite{Proenca2010}}

The imaging conditions engender acquired noisy artefacts that lead to severely degraded images, making iris segmentation a major issue. Having observed that existing iris segmentation methods tend to fail in these challenging conditions, authors present a segmentation method that can handle degraded images acquired in less constrained conditions. In this paper authors offer the following contributions: 1) to consider the sclera the most easily distinguishable part of the eye in degraded images, 2) to propose a new type of feature that measures the proportion of sclera in each direction and is fundamental in segmenting the iris, and 3) to run the entire procedure in deterministically linear time in respect to the size of the image, making the procedure suitable for real-time applications.

\paragraph{Iris Segmentation Methodology for Non-Cooperative Recognition~\cite{ProencaAlexandre2006}}

An overview of the iris image segmentation methodologies for biometric purposes is presented. The main focus is on the analysis of the ability of segmentation algorithms to process images with heterogeneous characteristics, simulating the dynamics of a non-cooperative environment. The accuracy of the four selected methodologies on the UBIRIS database is tested and, having concluded about their weak robustness when dealing with non-optimal images regarding focus, reflections, brightness or eyelid obstruction, the authors introduce a new and more robust iris image segmentation methodology. This new methodology could contribute to the aim of non-cooperative biometric iris recognition, where the ability to process this type of image is required.

\paragraph{Segmenting the Periocular Region using a Hierarchical Graphical Model Fed by Texture / Shape Information and Geometrical Constraints~\cite{ProencaSantos2014}}

Using the periocular region for biometric recognition is an interesting possibility: this area of the human body is highly discriminative among subjects and relatively stable in appearance. In this paper, the main idea is that improved solutions for defining the periocular region-of-interest and better pose / gaze estimates can be obtained by segment- ing (labelling) all the components in the periocular vicinity. Accordingly, authors describe an integrated algorithm for labelling the periocular region, that uses a unique model to discriminate between seven components in a single-shot: iris, sclera, eyelashes, eyebrows, hair, skin and glasses. The descibed solution fuses texture / shape descriptors and geometrical constraints to feed a two-layered graphical model (Markov Random Field), which energy minimization provides a robust solution against uncontrolled lighting conditions and variations in subjects pose and gaze.

\paragraph{Iris Recognition: A Method To Segment Visible Wavelength Iris Images Acquired On-The-Move and At-A-Distance~\cite{Proenca2008}}

The main point of this paper is to give a process suitable for the automatic segmentation of iris images captured at the visible wavelength, on-the-move and within a large range of image acquisition distances (between 4 and 8 meters). The experiments were performed on images of the UBIRIS.v2 database and show the robustness of the proposed method to handle the types of non-ideal images resultant of the aforementioned less constrained image acquisition conditions.

\paragraph{A Method for the Identification of Noisy Regions in Normalized Iris Images~\cite{ProencaAlexandre2006b}}

This paper proposes a new method for the identification of noisy regions in normalized iris images. Starting from a normalized and dimensionless iris image in the polar coordinate system, the goal is the classification of every pixel as ÓnoiseÓ or Ónot noiseÓ. This classification could be helpful in the posterior feature extraction or feature comparison stages regarding the construction of biometric iris signatures more robust to noise. They propose the extraction of 8 well known features for each pixel of the images followed by the classification through a neural network.

\paragraph{A Method for the Identification of Inaccuracies in the Pupil Segmentation~\cite{ProencaAlexandre2006c}}

This paper analyzes the relationship between the accuracy of the segmentation algorithm and the error rates of typical iris recognition systems. Authors selected 1000 images from the UBIRIS database that the segmentation algorithm can accurately segment and artificially introduced segmentation inaccuracies. authors repeated the recognition tests and concluded about the strong relationship between the errors in the pupil segmentation and the overall false reject rate. Based on this fact, they propose a method to identify these inaccuracies.

\section{General Works}

\paragraph{Visual Surveillance and Biometrics: Practices, Challenges, and Possibilities~\cite{BakshiTistarelli2019}}

Visual surveillance is the latest paradigm of monitoring public security through machine intelligence. It includes the use of visual data captured by IR sensors, cameras placed in car, corridors, traffic signals etc. Visual surveillance facilitates the classification of human behavior, crowd activity, and gesture analysis to achieve application-specific objectives.
Biometrics is the science of uniquely identifying or verifying an individual among a set of people by exploring the userÕs physiological or behavioral characteristics. Due to their ease of use in many application scenarios (including time attendance systems, border control, access control for high security, etc.), biometric systems are currently being introduced in many day-to-day activities.
Sometimes, algorithms developed for visual surveillance systems have been applied to biometric identification. Recently, several research efforts have been devoted to merge these two technologies especially for adverse and covert scenarios. This Special Section in IEEE Access  served as a cross-platform to cover the recent advancements at the intersection of Ôvisual surveillanceÕ and ÔbiometricsÕ.

\paragraph{IEEE Intelligent Systems: Trends and Controversies~\cite{ProencaNappi2018}}

Performing covert biometric recognition in surveillance environments has been regarded as a grand challenge, considering the adversity of the conditions where recognition should be carried out (e.g., poor resolution, bad lighting, off-pose and partially occluded data). This special issue compiles a group of approaches to this problem.

\paragraph{Biometric Recognition in Surveillance Scenarios: A Survey~\cite{NevesProenca2016}}

The interest in the security of individuals has increased in recent years.This increase has in turn led to much wider deployment of surveillance cameras worldwide, and con- sequently, automated surveillance systems research has received more attention from the scientific community than before. Concurrently, biometrics research has become more popular as well, and it is supported by the increasing number of approaches devised to address specific degradation factors of unconstrained environments. Despite these recent efforts, no automated surveillance system that performs reliable biometric recognition in such an environment has become available. Nevertheless, recent developments in human motion analysis and biometric recognition suggest that both can be combined to develop a fully automated system. As such, this paper reviews recent advances in both areas, with a special focus on surveillance scenarios. When compared to previous studies, authors highlight two distinct features, i.e., (1) put the emphasis is on approaches that are devised to work in unconstrained environments and surveillance scenarios; and (2) biometric recognition is the final goal of the surveillance system, as opposed to behavior analysis, anomaly detection or action recognition.

\paragraph{Fusing Vantage Point Trees and Linear Discriminants for Fast Feature Classification~\cite{ProencaNeves2017b}}

This paper describes a classification strategy that can be regarded as a more general form of nearest-neighbor classification. It fuses the concepts of nearest neighbor, linear discriminant and Vantage-Point trees, yielding an efficient indexing data structure and classification algorithm. In the learning phase, authors define a set of disjoint subspaces of reduced complexity that can be separated by linear discriminants, ending up with an ensemble of simple (weak) classifiers that work locally. In classification, the closest centroids to the query determine the set of classifiers considered, which responses are weighted. The algorithm was experimentally validated in datasets widely used in the field, attaining error rates that are favorably comparable to the state-of-the-art classification techniques. Lastly, the proposed solution has a set of interesting properties for a broad range of applications: 1) it is deterministic; 2) it classifies in time approximately logarithmic with respect to the size of the learning set, being far more efficient than nearest neighbor classification in terms of computational cost; and 3) it keeps the generalization ability of simple models.

\paragraph{Editorial of the Special Issue On the Recognition of Visible Wavelength Iris Images Captured At-a-distance and On-the-move~\cite{ProencaAlexandre2012}}

This special issue regards the recognition of degraded iris images acquired in visible wavelengths. During 2009 and 2010, the University of Beira Interior (Portugal) promoted two International evaluation initiatives about this subject, named Noisy Iris Challenge Evaluation (NICE) I and II. The first one focussed on the evaluation of iris segmentation strategies, considering that iris data acquired in visible wave- lengths (VW) usually has much higher level of detail than traditionally used near infra-red data (NIR), but also has many more noise arti- facts, including specular and diffuse reflections and shadows. Also, the spectral reflectance of the sclera is significantly higher in the VW than in the NIR and the spectral radiance of the iris with respect to the levels of its pigmentation varies much more significantly in the VW than in the NIR.

\paragraph{Introduction to the Special Issue on Unconstrained Biometrics: Advances and Trends~\cite{ProencaScharcanski2011}}

To date, no research effort has produced a machine able to autonomously and covertly perform reliable recognition of human beings. Perhaps, contrary to popular belief, such automata are confined to science fiction, although it is not hard to anticipate the potential impact that they would have in modern societies (e.g., forensics and surveillance). Some of the biological traits used to perform biometric recognition support contactless data acquisition and can be imaged covertly. Thus, at least theoretically, the subsequent biometric recognition procedure can be performed without subjectsÕ knowledge and in uncontrolled scenarios. This real-world scenario brings many challenges to the Pattern Recognition process, essentially due to poor quality of the acquired data. The feasibility of this type of recognition has received increasing attention and is of particular interest in visual surveillance, computer forensics, threat assessment, and other security areas.

\paragraph{Introduction to the Special Issue on the Segmentation of Visible Wavelength Iris Images Captured At-a-distance and On-the-move~\cite{ProencaAlexandre2010}}

Deployed iris recognition systems are mainly based on DaugmanÕs pioneering approach, and have proven their effectiveness in relatively constrained scenarios: operating in the near infra-red spectrum (NIR, 700-900 nm), at close acquisition distances and with stop-and-stare interfaces. However, the human iris supports contactless data acquisition, and it can - at least theoretically - be imaged covertly. The feasibility of covert iris recognition receives increasing attention and is of particular interest for forensic and security purposes. In this scope, one possibility is the use of visible wavelength light (VW) to perform image acquisition, although the use of this type of light can severely degrade the quality of the captured data. This is mainly due to the optical properties of the two molecules that constitute the pigment of the human iris: brown-black Eumelanin (over 90\%) and yellow-reddish Pheomelanin. Eumelanin has most of its radiative fluorescence under VW, which enables the capturing of a much higher level of detail, but also of many more noisy artefacts, including specular and diffuse reflections and shadows. Also, the spectral reflectance of the sclera is significantly higher in the VW than in the NIR and the spectral radiance of the iris in respect to the levels of its pigmentation varies much more significantly in the VW than in the NIR. Furthermore, traditional template- and boundary-based iris segmentation approaches will probably fail, due to difficulties in detecting edges or in fitting rigid shapes. All these reasons justify the need of specialized segmentation strategies and were the major motivations behind

\paragraph{Non-Cooperative Iris Recognition: Issues and trends~\cite{Proenca2011b}}

To date, no research effort has produced a machine able to covertly recognize human beings. Contrary to popular belief, such automata are confined to science fiction, although it is not hard to anticipate the potential impact that they would have in the security and safety of modern societies (forensics and surveillance). Among the research programs that pursuit such type of biometric recognition, previous initiatives sought to acquire data from moving subjects, at long distances and under uncontrolled lighting conditions. This real- world scenario brings many challenges to the Pattern Recognition process, essentially due to poor quality of the acquired data. Several programs now seek to increase the robustness to noise of each phase of the recognition process (detection, segmentation, normalization, encoding and matching). This paper addresses the feasibility of such extremely ambitious type of biometric recognition, discusses the major issues be- hind the development of this technology and points some directions for further improvements.

\paragraph{On the Feasibility of the Visible Wavelength, At-A-Distance and On-The-Move Iris Recognition~\cite{Proenca2009}}

The dramatic growth in practical applications for iris biometrics has been accompanied by relevant developments in the underlying algorithms and techniques. Among others, one active research area concerns about the development of iris recognition systems less constrained to users, either increasing the imaging distances, simplifying the acquisition protocols or the required lighting conditions. In this paper authors address the possibility of perform reliable recognition using visible wavelength images captured under high heterogeneous lighting conditions, with subjects at-a-distance (between 4 and 8 meters) and on-the-move. The feasibility of this extremely ambitious type of recognition is analyzed, its major obstacles and challenges discussed and some directions for forthcoming work pointed.

\section{Deep Learning}

\paragraph{Deep representations for cross-spectral ocular biometrics~\cite{ZanlorensiMenotti2019}}

One of the major challenges in ocular biometrics is the cross-spectral scenario, i.e., how to match images acquired in different wavelengths (typically visible (VIS) against near-infrared (NIR)). This article designs and extensively evaluates cross-spectral ocular verification methods, for both the closed and open-world settings, using well known deep learning representations based on the iris and periocular regions. Using as inputs the bounding boxes of non-normalized iris/periocular regions, authors fine-tune Convolutional Neural Network(CNN) models (based either on VGG16 or ResNet-50 architectures), originally trained for face recognition. Based on the experiments carried out in two publicly available cross-spectral ocular databases, authors report results for intra-spectral and cross-spectral scenarios, with the best performance being observed when fusing ResNet-50 deep representations from both the periocular and iris regions. When compared to the state-of-the-art, authors observed that the proposed solution consistently reduces the Equal Error Rate(EER) values by 90\% / 93\% / 96\% and 61\% / 77\% / 83\% on the cross-spectral scenario and in the PolyU Bi-spectral and Cross-eye-cross-spectral datasets. Lastly, authors evaluate the effect that the "deepness" factor of feature representations has in recognition effectiveness, and - based on a subjective analysis of the most problematic pairwise comparisons - authors point out further directions for this field of research.

\paragraph{"A Reminiscence of ÓMastermindÓ: Iris/Periocular Biometrics by ÓIn-SetÓ CNN Iterative Analysis~\cite{ProencaNeves2019}}

Convolutional neural networks (CNNs) have emerged as the most popular classification models in biometrics research. Under the discriminative paradigm of pattern recognition, CNNs are used typically in one of two ways: 1) verification mode (Óare samples from the same person?Ó), where pairs of images are provided to the network to distinguish between genuine and impostor instances; and 2) identification mode (Ówhom is this sample from?Ó), where appropriate feature representations that map images to identities are found. This paper postulates a novel mode for using CNNs in biometric identification, by learning models that answer to the question Óis the queryÕs identity among this set?Ó. The insight is a reminiscence of the classical Mastermind game: by iteratively analysing the network responses when multiple random samples of k gallery elements are compared to the query, authors obtain weakly correlated matching scores that - altogether - provide solid cues to infer the most likely identity. In this setting, identification is regarded as a variable selection and regularization problem, with sparse linear regression techniques being used to infer the matching probability with respect to each gallery identity. As main strength, this strategy is highly robust to outlier matching scores, which are known to be a primary error source in biometric recognition. Our experiments were carried out in full versions of two well known irises near-infrared (CASIA-IrisV4-Thousand) and periocular visible wavelength (UBIRIS.v2) datasets, and confirm that recognition performance can be solidly boosted-up by the proposed algorithm, when compared to the traditional working modes of CNNs in biometrics.

\paragraph{Deep-PRWIS: Periocular Recognition Without the Iris and Sclera Using Deep Learning Frameworks~\cite{ProencaNeves2018b}}

This work is based on a disruptive hypothesis for periocular biometrics: in visible-light data, the recognition performance is optimized when the components inside the ocular globe (the iris and the sclera) are simply discarded, and the recogniserÕs response is exclusively based in information from the surroundings of the eye. As major novelty, authors describe a processing chain based on convolution neural networks (CNNs) that defines the regions-of-interest in the input data that should be privileged in an implicit way, i.e., without masking out any areas in the learning/test samples. By using an ocular segmentation algorithm exclusively in the learning data, authors separate the ocular from the periocular parts. Then, authors produce a large set of Ómulti-classÓ artificial samples, by interchanging the periocular and ocular parts from different subjects. These samples are used for data augmentation purposes and feed the learning phase of the CNN, always considering as label the ID of the periocular part. This way, for every periocular region, the CNN receives multiple samples of different ocular classes, forcing it to conclude that such regions should not be considered in its response. During the test phase, samples are provided without any segmentation mask and the network naturally disregards the ocular components, which contributes for improvements in performance.

\paragraph{Pose Switch-Based Convolutional Neural Network for Clothing Analysis in Visual Surveillance Environments~\cite{AlirezazadehProenca2019}}

Recognizing pedestrian clothing types and styles in outdoor scenes and totally uncontrolled conditions is appealing to emerging applications such as security, intelligent customer profile analysis and computer-aided fashion design. Recognition of clothing categories from videos remains a challenge, mainly due to the poor data resolution and the data covariates that compromise the effectiveness of automated image analysis techniques (e.g., poses, shadows and partial occlusions). While state-of-the-art methods typically analyze clothing attributes without paying attention to variation of human poses, here authors claim for the importance of a feature representation derived from human poses to improve classification rate. Estimating the pose of pedestrians is important to fed guided features into recognizing system. In this paper, authors introduce pose switch-based convolutional neural network for recognizing the types of clothes of pedestrians, using data acquired in crowded urban environments. In particular, authors compare the effectiveness attained when using CNNs without respect to human poses variant, and assess the improvements in performance attained by pose feature extraction. The observed results enable us to conclude that pose information can improve the performance of clothing recognition system. authors focus on the key role of pose information in pedestrian clothing analysis, which can be employed as an interesting topic for further works.

\paragraph{FaceGenderID: Exploiting Gender Information in DCNNs Face Recognition Systems~\cite{BlasquezVera2019}}

This paper addresses the effect of gender as a covariate in face verification systems. Even though pre-trained models based on Deep Convolutional Neural Networks (DCNNs), such as VGG-Face or ResNet-50, achieve very high performance, they are trained on very large datasets comprising millions of images, which have biases regarding demographic aspects like the gender and the ethnicity among others. In this work, authors first analyse the separate performance of these state-of-the-art models for males and females. authors observe a gap between face verification performances obtained by both gender classes. These results suggest that features obtained by biased models are affected by the gender covariate. authors propose a gender-dependent training approach to improve the feature representation for both genders, and develop both: i) gender specific DCNNs models, and ii) a gender balanced DCNNs model. Our results show significant and consistent improvements in face verification performance for both genders, individually and in general with our proposed approach. Finally, authors announce the availability (at GitHub) of the FaceGenderID DCNNs models proposed in this work, which can support further experiments on this topic.

\paragraph{Segmentation-less and Non-holistic Deep-Learning Framework for Iris Recognition~\cite{ProencaNeves2019b}}

Driven by the pioneer iris biometrics approach, the most relevant recognition methods published over the years are Óphase-basedÓ, and segment/normalize the iris to obtain dimensionless representations of the data that attenuate the differences in scale, translation, rotation and pupillary dilation. In this paper authors present a recognition method that dispenses the iris segmentation, noise detection and normalization phases, and is agnostic to the levels of pupillary dilation, while maintaining state-of-the-art performance. Based on deep-learning classification models, authors anayze the displacements between biologically corresponding patches in pairs of iris images, to discriminate between genuine and impostor comparisons. Such corresponding patches are firstly learned in the normalized representations of the irises - the domain where they are optimally distinguishable - but are remapped into a segmentation-less polar coordinate system that uniquely requires iris detection. In recognition time, samples are only converted into this segmentation-less coordinate system, where matching is performed. 

\section{Recognition Works}

\paragraph{"A Leopard Cannot Change Its Spots": Improving Face Recognition Using 3D-based Caricatures~\cite{NevesProenca2018}}

Caricatures refer to a representation of a person in which the distinctive features are deliberately exaggerated, with several studies showing that humans perform better at recognizing people from caricatures than using original images. Inspired by this observation, this paper introduces the first fully automated caricature-based face recognition approach capable of working with data acquired in the wild. Our approach leverages the 3D face structure from a single 2D image and compares it to a reference model for obtaining a compact representation of face features deviations. This descriptor is subsequently deformed using a Õmea- sure locally, weight globallyÕ strategy to resemble the caricature drawing process. The deformed deviations are incorporated in the 3D model using the Laplacian mesh deformation algorithm, and the 2D face caricature image is obtained by projecting the deformed model in the original camera-view. To demonstrate the advantages of caricature-based face recognition, authors train the VGG-Face network from scratch using either original face images (baseline) or caricatured images, and use these models for extracting face descriptors from the LFW, IJB-A and MegaFace datasets. The experiments show an increase in the recognition accuracy when using caricatures rather than original images.

\paragraph{Joint Head Pose / Soft Label Estimation for Human Recognition In-The-Wild~\cite{ProencaMoreno2016}}

Soft biometrics have been emerging to complement other traits and are particularly useful for poor quality data. In this paper, authors propose an efficient algorithm to estimate human head poses and to infer soft biometric labels based on the 3D morphology of the human head. Starting by considering a set of pose hypotheses, authors use a learning set of head shapes synthesized from anthropometric surveys to derive a set of 3D head centroids that constitutes a metric space. Next, representing queries by sets of 2D head landmarks, authors use projective geometry techniques to rank efficiently the joint 3D head centroids / pose hypotheses according to their likelihood of matching each query. The rationale is that the most likely hypotheses are sufficiently close to the query, so a good solution can be found by convex energy minimization techniques. Once a solution has been found, the 3D head centroid and the query are assumed to have similar morphology, yielding the soft label. Our experiments point toward the usefulness of the proposed solution, which can improve the effectiveness of face recognizers and can also be used as a privacy-preserving solution for biometric recognition in public environments.

\paragraph{Robust periocular recognition by fusing sparse representations of color and geometry information~\cite{MorenoProenca2015}}

In this paper, authors propose a re-weighted elastic net (REN) model for biometric recognition. The new model is applied to data separated into geometric and color spatial components. The geometric information is extracted using a fast cartoon - texture decomposition model based on a dual formulation of the total variation norm allowing us to carry information about the overall geometry of images. Color components are defined using linear and nonlinear color spaces, namely the red-green-blue (RGB), chromaticity- brightness (CB) and hue-saturation-value (HSV). Next, according to a Bayesian fusion-scheme, sparse representations for classification purposes are obtained. The scheme is numerically solved using a gradient projection (GP) algorithm. In the empirical validation of the proposed model, authors have chosen the periocular region, which is an emerging trait known for its robustness against low quality data.

\paragraph{Aperiodic Feature Representation for Gait Recognition in Cross-view Scenarios for Unconstrained Biometrics~\cite{PadoleProenca2015}}

The state-of-the-art gait recognition algorithms require a gait cycle estimation before the feature extraction and are classified as periodic algorithms. Their effective- ness substantially decreases due to errors in detecting gait cycles, which are likely to occur in data acquired in non- controlled conditions. Hence, the main contributions of this paper are: (1) propose an aperiodic gait recognition strategy, where features are extracted without the concept of gait cycle, in case of multi-view scenario; (2) propose the fusion of the different feature subspaces of aperiodic feature representations at score level in cross-view scenarios. The experiments were performed with widely known CASIA Gait database B, which enabled us to draw the following major conclusions, (1) for multi-view scenarios, features extracted from gait sequences of varying length have as much discriminating power as traditional periodic features; (2) for cross-view scenarios, authors observed an average improvement of 22 \% over the error rates of state-of- the-art algorithms, due to the proposed fusion scheme.

\paragraph{Ocular Biometrics by Score-Level Fusion of Disparate Experts~\cite{Proenca2014b}}

The concept of periocular biometrics emerged to improve the robustness of iris recognition to degraded data. Being a relatively recent topic, most of the periocular recognition algorithms work in a holistic way, and apply a feature encoding / matching strategy without considering each biological component in the periocular area. This not only augments the correlation between the components in the resulting biometric signature, but also increases the sensitivity to particular data covariates. The main novelty in this paper is to propose a periocular recognition ensemble made of two disparate components: 1) one expert analyses the iris texture and exhaustively exploits the multi-spectral information in visible-light data; 2) another expert parameterises the shape of eyelids and defines a surrounding dimensionless region-of-interest, from where statistics of the eyelids, eyelashes and skin wrinkles / furrows are encoded. Both experts work on disjoint regions of the periocular area and meet three important properties: 1) they produce practically independent responses, which is behind the better performance of the ensemble when compared to the best individual recogniser; 2) they donÕt share particularly sensitivity to any image covariate, which accounts for augmenting the robustness against degraded data.

\paragraph{Face Recognition: Handling Data Misalignments Implicitly by Fusion of Sparse Representations~\cite{ProencaMoreno2014}}

Sparse representations for classification (SRC) are considered a relevant advance to the biometrics field, but are particularly sensitive to data misalignments. In previous studies, such misalignments were compensated for by finding appropriate geometric transforms between the elements in the dictionary and the query image, which is costly in terms of computational burden. This study describes an algorithm that compensates for data misalignments in SRC in an implicit way, that is, without finding/applying any geometric transform at every recognition attempt. The authorsÕ study is based on three concepts: (i) sparse representations; (ii) projections on orthogonal subspaces; and (iii) discriminant locality preserving with maximum margin projections. When compared with the classical SRC algorithm, apart from providing slightly better performance, the proposed method is much more robust against global/local data misalignments. In addition, it attains performance close to the state-of-the-art algorithms at a much lower computational cost, offering a potential solution for real- time scenarios and large-scale applications.

\paragraph{Periocular Biometrics: Constraining the EGM Algorithm to Biologically Plausible Distortions~\cite{ProencaMoreno13}}

In biometrics research, the periocular region has been regarded as an interesting trade-off between the face and the iris, particularly in unconstrained data acquisition setups. As in other biometric traits, the current challenge is the development of more robust recognition algorithms. Having investigated the suitability of the Ôelastic graph matchingÕ (EGM) algorithm to handle non- linear distortions in the periocular region because of facial expressions, the authors observed that vertices locations often not correspond to displacements in the biological tissue. Hence, they propose a Ôglobally coherentÕ variant of EGM (GC-EGM) that avoids sudden local angular movements of vertices while maintains the ability to faithfully model non-linear distortions. Two main adaptations were carried out: (i) a new term for measuring vertices similarity and (ii) a new term in the edges-cost function penalises changes in orientation between the model and test graphs.

\paragraph{Compensating for Pose and Illumination in Unconstrained Periocular Biometrics~\cite{PadoleProenca2013}}

In the context of less constrained biometrics recognition, the use of information from the vicinity of the eyes (periocular) is considered with high potential and motivated several recent proposals. In this paper, authors focus on two factors that are known to degrade the performance of periocular recognition: varying illumination conditions and subjects pose. Hence, this paper has three major purposes: 1) describe the decreases in performance due to varying illumination and subjects poses; 2) propose two techniques to improve the robustness to these factors; 3) announce the availability of an annotated dataset of periocular data (UBIPosePr), where poses vary in regular intervals, turning it especially suitable to assess the effects of misalignments between camera and subjects in periocular recognition.

\paragraph{Fusing Color and Shape Descriptors in the Recognition of Degraded Iris Images Acquired at Visible Wavelength~\cite{ProencaSantos20212}}

Despite the substantial research into the development of covert iris recognition technologies, no machine to date has been able to reliably perform recognition of human beings in real-world data. This limitation is especially evident in the application of such technology to large-scale identification scenarios, which demand extremely low error rates to avoid frequent false alarms. Most previously published works have used intensity data and performed multi-scale analysis to achieve recognition, obtaining encouraging performance values that are nevertheless far from desirable. This paper presents two key innovations. (1) A recognition scheme is proposed based on techniques that are substantially different from those traditionally used, starting with the dynamic partition of the noise-free iris into disjoint regions from which MPEG-7 color and shape descriptors are extracted. (2) The minimal levels of linear correlation between the outputs produced by the proposed strategy and other state-of-the-art techniques suggest that the fusion of both recognition techniques significantly improve performance, which is regarded as a positive step towards the development of extremely ambitious types of biometric recognition.

\paragraph{An Iris Recognition Approach Through Structural Pattern Analysis Methods~\cite{Proenca2010b}}

Continuous efforts have been made in searching for robust and effective iris cod- ing methods, since DaugmanÕs pioneering work on iris recognition was published. Proposed algorithms follow the statistical pattern recognition paradigm and encode the iris texture in- formation through phase, zero-crossing or texture-analysis based methods. In this paper authors propose an iris recognition algorithm that follows the structural (syntactic) pattern recognition paradigm, which can be advantageous essentially for the purposes of description and of the human-perception of the systemÕs functioning. Our experiments, that were performed on two widely used iris image databases (CASIA.v3 and ICE), show that the proposed iris structure provides enough discriminating information to enable accurate biometric recognition, while maintains the advantages intrinsic to structural pattern recognition systems.

\paragraph{Toward Non-Cooperative Iris Recognition: A Classification Approach Using Multiple Signatures~\cite{ProencaAlexandre2007}}

This paper focus on noncooperative iris recognition, i.e., the capture of iris images at large distances, under less controlled lighting conditions, and without active participation of the subjects. This increases the probability of capturing very heterogeneous images (regarding focus, contrast, or brightness) and with several noise factors (iris obstructions and reflections). Current iris recognition systems are unable to deal with noisy data and substantially increase their error rates, especially the false rejections, in these conditions. authors propose an iris classification method that divides the segmented and normalized iris image into six regions, makes an independent feature extraction and comparison for each region, and combines each of the dissimilarity values through a classification rule.

\paragraph{Region-Based CNNs for Pedestrian Gender Recognition in Visual Surveillance Environments~\cite{YaghoubiProenca2019}}

Inferring soft biometric labels in totally uncontrolled outdoor environments, such as surveillance scenarios, remains a challenge due to the low resolution of data and its covariates that might seriously compromise performance (e.g., occlusions and subjects pose). In this kind of data, even state-of-the-art deep-learning frameworks (such as ResNet) working in a holistic way, attain relatively poor performance, which was the main motivation for the work described in this paper. In particular, having noticed the main effect of the subjects' ÒposeÓ factor, in this paper authors describe a method that uses the body keypoints to estimate the subjects pose and define a set of regions of interest (e.g., head, torso, and legs). This information is used to learn appropriate classification models, specialized in different poses/body parts, which contributes to solid improvements in performance. This conclusion is supported by the experiments authors conducted in multiple real-world outdoor scenarios, using the data acquired from advertising panels placed in crowded urban environments.

\paragraph{IRINA: Iris Recognition (even) in Inacurately Segmented Data~\cite{ProencaNeves2017c}}

The effectiveness of current iris recognition systems de- pends on the accurate segmentation and parameterisation of the iris boundaries, as failures at this point misalign the coefficients of the biometric signatures. This paper de- scribes IRINA, an algorithm for Iris Recognition that is ro- bust against INAccurately segmented samples, which makes it a good candidate to work in poor-quality data. The pro- cess is based in the concept of ÓcorrespondingÓ patch be- tween pairs of images, that is used to estimate the posterior probabilities that patches regard the same biological region, even in case of segmentation errors and non-linear texture deformations. Such information enables to infer a free-form deformation field (2D registration vectors) between images, whose first and second-order statistics provide effective bio- metric discriminating power.

\paragraph{Quis-Campi: Extending In The Wild Biometric Recognition to Surveillance Environments~\cite{SantosProenca2015b}}

Efforts in biometrics are being held into extending robust recognition techniques to in the wild scenarios. Nonetheless, and despite being a very attractive goal, human identification in the surveillance con- text remains an open problem. In this paper, authors introduce a novel biometric system  Quis-Campi  that effectively bridges the gap between surveillance and biometric recognition while having a minimum amount of operational restrictions. authors propose a fully automated surveillance sys- tem for human recognition purposes, attained by combining human detection and tracking, further enhanced by a PTZ camera that delivers data with enough quality to perform biometric recognition. Along with the system concept, implementation details for both hardware and software modules are provided, as well as preliminary results over a real scenario

\paragraph{Robust Periocular Recognition by Fusing Local to Holistic Sparse Representations~\cite{MorenoProenca2013}}

Sparse representations have been advocated as a relevant advance in biometrics research. In this paper authors propose a new algorithm for fusion at the data level of sparse representations, each one obtained from image patches. The main novelties are two-fold: 1) a dictionary fusion scheme is formalised, using the l1minimization with the gradient projection method; 2) the proposed representation and classification method does not require the non-overlapping condition of image patches from where individual dictionaries are obtained.
In the experiments, authors focused in the recognition of periocular images and obtained independent dictionaries for the eye, eyebrow and skin regions, that were subsequently fused. Results obtained in the publicly available UBIRIS.v2 data set show consistent improvements in the recognition effectiveness when compared to state-of-the-art related representation and classification techniques.

\paragraph{Periocular Biometrics: An Emerging technology for Unconstrained Scenarios~\cite{SantosProenca2013}}

The periocular region has recently emerged as a promising trait for unconstrained biometric recognition, specially on cases where neither the iris and a full facial picture can be obtained. Previous studies concluded that the regions in the vicinity of the human eye - the periocular region- have surprisingly high discriminating ability between individuals, are relatively permanent and easily acquired at large distances. Hence, growing attention has been paid to periocular recognition methods, on the performance levels they are able to achieve, and on the correlation of the responses given by other. This work overviews the most relevant research works in the scope of periocular recognition: summarizes the developed methods, and enumerates the current issues, providing a comparative overview. For contextualization, a brief overview of the biometric field is also given.

\paragraph{Iris Recognition: A Method to Increase the Robustness to Noisy Imaging Environments Through the Selection of the Higher Discriminating Features~\cite{ProencaAlexandre2007c}}

Continuous efforts have been made in searching for robust and effective iris coding methods, since DaugmanÕs pioneering work on iris recognition was published. However, due to lack of robustness, the error rates of iris recognition systems significantly increase when images contain large portions of noise (reflections and iris obstructions), resultant from less constrained imaging conditions. Current iris encoding and matching proposals do not take into account the specific lighting conditions of the imaging environment, decreasing their adaptability to such dynamics conditions. In this paper authors propose a method that, through a learning stage, takes into account the typical noisy regions propitiated by the imaging environment to select the higher discriminating features.

\paragraph{Iris Recognition: An Entropy-Based Coding Strategy Robust to Noisy Imaging Environments~\cite{ProencaAlexandre2007b}}

The iris is currently accepted as one of the most accurate traits for biometric purposes. However, for the sake of accuracy, iris recognition systems rely on good quality images and significantly deteriorate their results when images contain large noisy regions, either due to iris obstructions (eyelids or eyelashes) or reflections (specular or lighting). In this paper authors propose an entropy-based iris coding strategy that constructs an unidimensional signal from overlapped angular patches of normalized iris images. Further, in the comparison between biometric signatures authors exclusively take into account signaturesÕ segments of varying dimension. The hope is to avoid the comparison between components corrupted by noise and achieve accurate recognition, even on highly noisy images. Our experiments were performed in three widely used iris image databases (third version of CASIA, ICE and UBIRIS) and led us to observe that our proposal significantly decreases the error rates in the recognition of noisy iris images.

\paragraph{A Structural Pattern Analysis Approach to Iris Recognition~\cite{Proenca2007}}

Continuous efforts have been made in searching for robust and effective iris cod- ing methods, since DaugmanÕs pioneering work on iris recognition was published. Proposed algorithms follow the statistical pattern recognition paradigm and encode the iris texture in- formation through phase, zero-crossing or texture-analysis based methods. In this paper authors propose an iris recognition algorithm that follows the structural (syntactic) pattern recognition paradigm, which can be advantageous essentially for the purposes of description and of the human-perception of the systemÕs functioning. Our experiments, that were performed on two widely used iris image databases (CASIA.v3 and ICE), show that the proposed iris structure provides enough discriminating information to enable accurate biometric recognition, while maintains the advantages intrinsic to structural pattern recognition systems.

\paragraph{Iris Recognition: Measuring FeatureÕs Quality for the Feature Selection in Unconstrained Image Capture Environments~\cite{ProencaAlexandre2006d}}

Iris recognition has been used for several purposes. However, current iris recognition systems are unable to deal with noisy data and substantially increase their error rates, specially the false rejections, in these conditions. Several proposals have been made to access image quality and to identify noisy regions in iris images. In this paper authors propose a method that measures the quality of each feature of the biometric signature and takes account into this information to constraint the comparable features and obtain the similarity between iris signatures. Experiments led us to conclude that this method significantly decreases the error rates in the recognition of noisy iris images, resultant from capturing in less constrained environments.

\section{Empirical Works}

\paragraph{Experiments with Ocular Biometric Datasets: A PractitionerÕs Guideline~\cite{AkhtarProenca2018}}

Ocular biometrics is the imaging and use of features extracted from the eyes regions for personal recognition. Ocular biometrics is a promising research field owing to factors such as recognition at a distance and suitability for recognition with regular RGB cameras, especially in visible spectrum on mobile devices. To ensure that ocular biometric academic researches have a positive impact on future technological developments, this paper provides a review of ocular databases available in litera- ture, diversities among these databases, design and parameters consideration issues during acquisition of database and selection of appropriate database for experimentation. Open issues and future research directions are also discussed to identify the path forward.

\paragraph{Insights into the results of MICHE I - Mobile Iris CHallenge Evaluation~\cite{MarsicoProenca2017}}

Mobile biometrics technologies are nowadays the new frontier for secure use of data and services, and are considered particularly important due to the massive use of handheld devices in the entire world. Among the biometric traits with potential to be used in mobile settings, the iris/ocular region is a natural candidate, even considering that further advances in the technology are required to meet the operational requirements of such ambitious environments. Aiming at promoting these advances, authors organized the Mobile Iris Challenge Evaluation (MICHE)-I contest. This paper presents a comparison of the performance of the participant methods by various Figures of Merit (FoMs). A particular attention is devoted to the identification of the image covariates that are likely to cause a decrease in the performance levels of the compared algorithms. Among these factors, interoperability among different devices plays an important role. The methods (or parts of them) implemented by the analyzed approaches are classified into segmentation (S), which was the main target of MICHE-I, and recognition (R). The paper reports both the results observed for either S or R, and also for different recombinations (S+R) of such methods. Last but not least, authors also present the results obtained by multi-classifier strategies.

\paragraph{Results from MICHE II - Mobile Iris CHallenge Evaluation II~\cite{MarsicoProenca2017b}}

Mobile biometrics represent the new frontier of authentication. The most appealing feature of mobile devices is the wide availability and the presence of more and more reliable sensors for capturing bio- metric traits, e.g., cameras and accelerometers. Moreover, they more and more often store personal and sensitive data, that need to be protected. Doing this on the same device using biometrics to enforce security seems a natural solution. This makes this research topic attracting and generally promising. However, the growing interest for related applications is counterbalanced by still present limitations, especially for some traits. Acquisition and computation resources are nowadays widely available, but they are not al- ways sufficient to allow a reliable recognition result. Most of all, the way capture is expected to be carried out, i.e., by the user him/herself in uncontrolled conditions and without an expert assistance, can heavily affect the quality of samples and, as a consequence, the accuracy of recognition. Among the biometric traits raising the interest of researchers, iris plays an important role. Mobile Iris CHallenge Evaluation II (MICHE II) competition provided a testbed to assess the progress of mobile iris recognition, as well as its limitations still to overcome. This paper presents the results of the competition and the analysis of achieved performance, that takes into account both proposals submitted for the competition section launched at the 2016 edition of the International Conference on Pattern Recognition (ICPR), as well as proposals submitted for this special issue.

\paragraph{Periocular Recognition: How Much Facial Expressions Affect Performance?~\cite{BarrosoProenca2015}}

Using information near the human eye to per- form biometric recognition has been gaining popularity. Previous works in this area, designated periocular recognition, show remarkably low error rates and particularly high robustness when data are acquired under less con- trolled conditions. In this field, one factor that remains to be studied is the effect of facial expressions on recognition performance, as expressions change the textural/shape information inside the periocular region. authors have collected a multisession dataset whose single variation is the subjectsÕ facial expressions and analyzed the corresponding variations in performance, using the state-of-the-art peri- ocular recognition strategy. The effectiveness attained by different strategies to handle the effects of facial expressions was compared: (1) single-sample enrollment; (2) multisample enrollment, and (3) multisample enrollment with facial expression recognition, with results also vali- dated in the well-known Cohn-Kanade AU-Coded Expression dataset. Finally, the role of each type of facial expression in the biometrics menagerie effect is discussed.

\paragraph{Mobile Iris CHallenge Evaluation II: results from the ICPR competition.~\cite{CastrillonProenca2016}}

The growing interest for mobile biometrics stems from the increasing need to secure personal data and services, which are often stored or accessed from there. Modern user mobile devices, with acquisition and com- putation resources to support related operations, are nowadays widely available. This makes this research topic very attracting and promising. Iris recognition plays a major role in this scenario. However, mo- bile biometrics still suffer from some hindering fac- tors. The resolution of captured images and the com- putational power are not comparable to desktop sys- tems yet. Furthermore, the acquisition setting is gener- ally uncontrolled, with users who are not that expert to autonomously generate biometric samples of sufficient quality. Mobile Iris CHallenge Evaluation aims at pro- viding a testbed to assess the progress of mobile iris recognition, and to evaluate the extent of its present lim- itations. This paper presents the results of the compe- tition launched at the 2016 edition of the International Conference on Pattern Recognition (ICPR)

\paragraph{Iris Recognition: What's Beyond Bit Fragility?~\cite{Proenca2015}}

The concept of fragility of some bits in the iris codes regards exclusively their within-class variation, i.e., the probability that they take different values in templates computed from different images of the same iris. This paper extends that concept, by noticing that a similar phenomenon occurs for the between-classes comparisons, i.e., some bits have higher probability than others of assuming a predominant value, which was observed for near-infrared and (in a more evident way) for visible wavelength data. Accordingly, authors propose a new measure (bit discriminability) that takes into account both the within-class and between-classes variabilities, and has roots in the Fisher discriminant. Based on the bit discriminability, authors compare the usefulness of the different regions of the iris for biometric recognition, with respect to multi-spectral data and to different filters parameterizations. Finally, authors measure the amount of information lost in codes quantization, which gives insight to further research on iris matching strategies that consider both phase and magnitude. Albeit augmenting the computational burden of recognition, such kind of strategies will consistently improve performance, particularly in poor-quality data.

\paragraph{Toward Covert Iris Biometric Recognition: Experimental Results From the NICE Contests~\cite{ProencaAlexandre2012b}}

This paper announces and discusses the experimental results from the Noisy Iris Challenge Evaluation (NICE), an iris biometric evaluation initiative that received worldwide participation and whose main innovation is the use of heavily degraded data acquired in the visible wavelength and uncontrolled setups, with subjects moving and at widely varying distances. The NICE contest included two separate phases: 1) the NICE.I evaluated iris segmentation and noise detection techniques and 2) the NICE:II evaluated encoding and matching strategies for biometric signatures. Further, authors give the performance values observed when fusing recognition methods at the score level, which was observed to outperform any isolated recognition strategy. These results provide an objective estimate of the potential of such recognition systems and should be regarded as reference values for further improvements of this technology, which - if successful - may significantly broaden the applicability of iris biometric systems to domains where the subjects cannot be expected to cooperate.

\paragraph{Iris Recognition: Analysis of the Error Rates Regarding the Accuracy of the Segmentation Stage~\cite{ProencaAlexandre2010c}}

Iris recognition has been widely used in several scenarios with very satisfactory results. As it is one of the earliest stages, the image segmentation is in the basis of the process and plays a crucial role in the success of the recognition task. In this paper authors analyze the relationship between the accuracy of the iris segmentation process and the error rates of three typical iris recognition methods. authors selected 5000 images of the UBIRIS, CASIA and ICE databases that the used segmentation algorithm can accurately segment and artificially simulated four types of segmentation inaccuracies. The obtained results allowed us to con- clude about a strong relationship between translational segmentation inaccuracies - that lead to errors in phase - and the recognition error rates.

\paragraph{ICB-RW 2016: International Challenge on Biometric Recognition in the Wild~\cite{NevesProenca2016b}}

Biometric recognition in totally wild conditions, such as the observed in visual surveillance scenarios has not been achieved yet. The ICB-RW competition was promoted to support this endeavor, being the first biometric challenge carried out in data that realistically result from surveillance scenarios. The competition relied on an innovative master- slave surveillance system for the acquisition of face imagery at-a-distance and on-the-move. This paper describes the competition details and reports the performance achieved by the participants algorithms.

\paragraph{Evaluation of Background Subtraction Algorithms for Human Visual Surveillance~\cite{NevesProenca2015d}}

The fully automated surveillance of human be- ings remains an open problem, particularly for in-the-wild scenarios, i.e., for complex backgrounds and under uncon- trolled lighting conditions. Background Subtraction (BGS) is typically the first phase of the processing chain of such type of systems and holds the feasibility of all the subsequent phases. Hence, it is particularly important to perceive the relative effectiveness of BGS, with respect to the kind of environment. This paper gives an objective evaluation of the state-of-the-art BGS algorithms on unconstrained outdoor environments. When compared to similar published works, the major novelties are two-fold: 1) the focus is put on scenes populated by human beings; and 2) an objective measure of the wildness of environments is proposed, that strongly correlates to BGS performance, and enables to perceive the algorithmsÕ robustness with respect to the environment complexity. As main conclusions, authors observed that the SOBS algorithm outperforms the remaining methods. Nevertheless, its performance leads to conclude that BGS in unconstrained environments is still an open problem.

\paragraph{Creating Synthetic IrisCodes to Feed Biometrics Experiments~\cite{ProencaNeves2017d}}

The collection of iris data suitable to be used in experiments is difficult, mainly due to two factors: 1) the time spent by volunteers in the acquisition process; and 2) security / privacy concerns of volunteers. Even though there are methods to create images of artificial irises, there is no method exclusively focused in the synthesis of the iris biometric signatures (IrisCodes). In experiments related with some phases of the biometric recognition process (e.g., indexing / retrieval), a large number of signatures is required for proper evaluation, which, in case of real data, is extremely hard to obtain. Hence, this paper describes a stochastic method to synthesize IrisCodes, based on the notion of data correlation. These artificial signatures can be used to feed experiments on iris recognition, namely on the iris matching, indexing and retrieval phases. authors experimentally confirmed that both the genuine and impostor distributions obtained on the artificial data closely resemble the values obtained in data sets of real irises. Finally, another interesting feature is that the method is easily parametrized to mimic IrisCodes extracted from data of varying levels of quality, i.e., ranging from data acquired in high controlled to unconstrained environments.

\paragraph{Facial Expressions: Discriminability of Facial Regions and Relationship to Biometrics Recognition~\cite{BarrosoProenca2013}}

Facial expressions result from movements of mus- cular action units, in response to internal emotion states or perceptions, and it has been shown that they decrease the performance of face-based biometric recognition techniques. This paper focuses in the recognition of facial expressions and has the following purposes: 1) confirm the suitability of using dense image descriptors widely known in biometrics research (e.g., local binary patterns and histograms of oriented gradients) to recognize facial expressions; 2) compare the effectiveness attained when using different regions of the face to recognize expressions; 3) compare the effectiveness attained when the identity of subjects is known / unknown, before attempting to recognize their facial expressions.

\paragraph{Periocular Recognition: Analysis of Performance Degradation Factors~\cite{PadoleProenca2012}}

Among the available biometric traits such as face, iris and fingerprint, there is an active research being carried out in the direction of unconstrained biometrics. Periocular recognition has proved its effectiveness and is regarded as complementary to iris recognition. The main objectives of this paper are three-fold: 1) to announce the availability of periocular dataset, which has a variability in terms of scale change (due to camera-subject distance), pose variation and non-uniform illumination; 2) to investigate the performance of periocular recognition methods with the presence of various degradation factors; 3) propose a new initialization strategy for the definition of the periocular region-of- interest (ROI), based on the geometric mean of eye corners. Our experiments confirm that performance can be consis- tently improved by this initialization method, when com- pared to the classical technique.

\paragraph{Iris Recognition: Preliminary Assessment about the Discriminating Capacity of Visible Wavelength Data~\cite{SantosProenca2010b}}

The human iris supports contactless data acquisition and can be imaged covertly. These factors give raise to the possibility of performing biometric recognition procedure with- out subjectsÕ knowledge and in uncontrolled data acquisition scenarios. The feasibility of this type of recognition has been receiving increasing attention, as is of particular interest in visual surveillance, computer forensics, threat assessment, and other security areas. In this paper authors stress the role played by the spectrum of the visible light used in the acquisition process and assess the discriminating iris patterns that are likely to be acquired according to three factors: type of illuminant, itÕs luminance, and levels of iris pigmentation. Our goal is to perceive and quantify the conditions that appear to enable the biometric recognition process with enough confidence

\paragraph{Iris Recognition: Analysing the Distribution of the Iriscodes Concordant Bits~\cite{SantosProenca2010}}

The growth in practical applications for iris bio- metrics has been accompanied by relevant developments in the underlying algorithms and techniques. Efforts are being made to minimize the tradeoff between the recognition error rates and data quality, acquired in the visible wavelength, in less controlled environments, over simplified acquisition protocols and varying lighting conditions. This paper presents an approach that can be regarded as an extension to the widely known DaugmanÕs method. Its basis is the analysis of the distribution of the concordant bits when matching iriscodes on both the spatial and frequency domains. Our experiments show that this method is able to improve the recognition performance over images captured in less constrained acquisition setups and protocols. Such conclusion was drawn upon trials conducted for multiple datasets.

\paragraph{Biometric Recognition: When Is Evidence Fusion Advantageous?~\cite{Proenca2009b}}

Having assessed the performance gains due to evidence fusion, previous works reported contradictory conclusions. For some, a consistent improvement is achieved, while others state that the fusion of a stronger and a weaker biometric expert tends to produce worst results than if the best expert was used individually. The main contribution of this paper is to assess when improvements in performance are actually achieved, regarding the individual performance of each expert. Starting from readily satisfied assumptions about the score distributions generated by a biometric system, authors predict the performance of each of the individual experts and of the fused system. Then, authors conclude about the performance gains in fusing evidence from multiple sources. Also, authors parameterize an empirically obtained relationship between the individual performance of the fused experts that contributes to decide whether evidence fusion techniques are advantageous or not.

\paragraph{On the Role of Interpolation in the Normalization of Non-Ideal Visible Wavelength Iris Images~\cite{SantosProenca2009}}

The growth in practical applications for iris bio- metrics has been accompanied by relevant developments in the underlying algorithms and techniques. Along with the research focused on near-infrared (NIR) cooperatively captured images, efforts are being made to minimize the trade-off between the quality of the captured data and the recognition accuracy on less constrained environments, where images are obtained at the visible wavelength, at increased distances, over simplified protocols and adverse lightning. This paper addresses the effect of the interpolation method, used in the iris normalization stage, in the overall recognition error rates. This effect is stressed for systems operating under less constrained image acquisition setups and protocols, due to higher variations in the amounts of captured data. Our experiments led us to conclude that the utility of the image interpolating methods is directly corresponding to the levels of noise that images contain.

\paragraph{The NICE.I: Noisy Iris Challenge Evaluation - Part I~\cite{ProencaAlexandre2007d}}

This paper gives an overview of the NICE.I : Noisy Iris Challenge Evaluation - Part I contest. This contest differs from others in two fundamental points. First, instead of the complete iris recognition process, it exclusively evaluates the iris segmentation and noise detection stages, allowing the independent evaluation of one of the main recognition error sources. Second, it operates on highly noisy images that were captured to simulate less constrained imaging environments and constitute the second version of the UBIRIS database (UBIRIS.v2).

\paragraph{Iris Recognition: An Analysis of the Aliasing Problem in the Iris Normalization Stage~\cite{ProencaAlexandre2006e}}

Iris recognition has been increasingly used with very satisfactory results. Presently, the challenge consists in unconstrain the image capturing conditions and enable its application to domains where the subjectsÕ cooperation is not expectable (e.g. criminal/terrorist seek, missing children). In this type of use, due to variations in the image capturing distance and in the lighting conditions that determine the size of the subjectsÕ pupil, the area correspondent to the iris in the captured images will be highly varying too. In order to compensate this variation, common iris recognition proposals translate the segmented iris image to a double dimensionless pseudo-polar coordinate system, in a process known as the normalization stage, which can be regarded as a sampling of the original data with the inherent possibility of aliasing. In this paper authors analyze the relationship between the size of the captured iris image and the overall recognitionÕs accuracy. Further, authors identify the threshold for the sampling rate of the iris normalization process above which the error rates significantly increase.

\section{Other Works}

\paragraph{Quality Assessment of Degraded Iris Images Acquired in the Visible Wavelength~\cite{Proenca2011}}

Data quality assessment is a key issue, in order to broad the applicability of iris biometrics to unconstrained imaging conditions. Previous research efforts sought to use visible wavelength (VW) light imagery to acquire data at significantly larger distances than usual and on moving subjects, which makes this real world data notoriously different from the acquired in the near infra-red (NIR) setup. Having empirically observed that published strategies to assess iris image quality do not handle the specificity of such data, this paper proposes a method to assess the quality of VW iris samples captured in unconstrained conditions, according to the factors that are known to determine the quality of iris biometric data: focus, motion, angle, occlusions, area, pupillary dilation and levels of iris pigmentation. The key insight is to use the output of the segmentation phase in each assessment, which permits to handle severely degraded samples that are likely to result of such imaging setup. Also, our experiments point that the given method improves the effectiveness of VW iris recognition, by avoiding that poor quality samples are considered in the recognition process.

\paragraph{Exploiting Data Redundancy for Error Detection in Degraded Biometric Signatures Resulting From in the Wild Environments~\cite{NevesProenca2017}}

An error-correcting code (ECC) is a process of adding redundant data to a message, such that it can be recovered by a receiver even if a number of errors are introduced in transmission. Inspired by the principles of ECC, authors introduce a method capable of detecting degraded features in biometric signatures by exploiting feature correlation. The main novelty is that, unlike existing biometric cryptosystems, the proposed method works directly on the biometric signature. Our approach performs a redundancy analysis of non-degraded data to build an undirected graphical model (Markov Random Field), whose energy minimization determines the sequence of degraded components of the biometric sample. Experiments carried out in different biometric traits ascertain the improvements attained when disregarding degraded features during the matching phase. Also, authors stress that the proposed method is general enough to work in different classification methods, such as CNNs.

\paragraph{Multimodal Ocular Biometrics Approach: A Feasibility Study~\cite{KomogortsevProenca2012}}

Growing efforts have been concentrated on the development of alternative biometric recognition strategies, the intended goal to increase the accuracy and counterfeit-resistance of existing systems without increased cost. In this paper, authors propose and evaluate a novel biometric approach using three fundamentally different traits captured by the same camera sensor. Considered traits include: 1) the internal, non-visible, anatomical properties of the human eye, represented by Oculomotor Plant Characteristics (OPC); 2) the visual attention strategies employed by the brain, represented by Complex Eye Movement pat- terns (CEM); and, 3) the unique physical structure of the iris. Our experiments, performed using a low-cost web camera, indicate that the combined ocular traits improve the accuracy of the resulting system. As a result, the combined ocular traits have the potential to enhance the accuracy and counterfeit-resistance of existing and future biometric systems.

\paragraph{A Robust Eye-Corner Detection Method for Real-World Data~\cite{SantosProenca2011}}

Corner detection has been motivating several research works and is particularly important in different computer vision tasks, acting as basis for further image understanding stages. Particularly, the detection of eye-corners in facial images is relevant for domains such as biometric systems and assisted-driving systems. Having empirically evaluated the state-of-the-art of eye-corner detection proposals, authors observed that they only achieve satisfactory results when dealing with good quality data. Hence, in this paper authors describe an eye- corner detection method with particular focus on robustness, i.e., the suitability to deal with degraded data, toward the applicability in real-world conditions. Our experiments show that the proposed method outperforms others either in noise- free and degraded data (blurred, rotated and with significant variations in scale), which is regarded as the main achievement.

\paragraph{Combining Rectangular and Triangular Image Regions to Perform Real-Time Face Detection~\cite{ProencaFilipe2008}}

Nowadays, face detection techniques assume grow- ing relevance in a wide range of applications (e.g., bio- metrics and automatic surveillance)and constitute a pre- requisite of many image processing stages. Among a large number of published approaches, one of the most relevant is the method proposed by Viola and Jones to perform real-time face detection through a cascade schema of weak classifiers that act together to compose a strong and robust classifier. This method was the basis of our work and motivated the key contributions given in this paper. At first, based on the computer graphics concept of Ótriangle meshÓ authors propose the notion of Ótriangular integral featureÓ to describe and model face properties. Also, authors show results of our face detection experiments that point to an increase of the detection accuracy when the triangular features are mixed with the rectangular in the candidate feature set, which is considered an achievement. Also, it should be stressed that this optimization is obtained without any relevant increase in the computational requirements, either spacial or temporal, of the detection method.

\paragraph{Iris Biometrics: Indexing and Retrieving Heavily Degraded Data~\cite{Proenca2013}}

Most of the methods to index iris biometric signa- tures were designed for decision environments with a clear separation between genuine and impostor matching scores. However, in case of less controlled data acquisition, images will be degraded and the decision environments poorly separated. This paper proposes an indexing / retrieval method for degraded images and operates at the code level, making it compatible with different feature encoding strategies. Gallery codes are decomposed at multiple scales, and according to their most reliable components at each scale, the position in an n-ary tree determined. In retrieval, the probe is decomposed similarly, and the distances to multi-scale centroids are used to penalize paths in the tree. At the end, only a subset of the branches is traversed up to the last level. When compared to related strategies, the proposed method outperforms them on degraded data, particularly in the performance range most important for biometrics (hit rates above 0.95). Finally, according to the computational cost of the retrieval phase, the number of enrolled identities above which indexing is computationally cheaper than an exhaustive search is determined.

\section{Conclusions}

The development of biometric recognition solutions able to work in visual surveillance conditions, i.e., in unconstrained data acquisition conditions and under covert protocols has been motivating growing  efforts  from the research community.  Among the various laboratories, schools and research institutes concerned  about this problem, the SOCIA: Soft Computing and Image Analysis Lab., of the University of Beira Interior, Portugal, has been among the most active in pursuing disruptive solutions for obtaining such extremely ambitious kind of automata. This report summarised the research works published by elements of the SOCIA Lab. in the last decade in the scope of biometric recognition in unconstrained conditions. Hopefully, it can be used as a reference for someone entering in this extremely challenging research topic.

{\small
\bibliographystyle{ieee}

}

\end{document}